\documentclass[10pt,twocolumn]{article}

\usepackage[a4paper,top=0.72in,bottom=0.78in,left=0.72in,right=0.72in,columnsep=0.25in]{geometry}
\usepackage[round,authoryear]{natbib}
\usepackage[hidelinks]{hyperref}
\hypersetup{
  pdftitle={Toward Real-Time Sentence-Level Sign Language Translation},
  pdfauthor={Thanh-Hoang Nguyen Doan}
}
\usepackage{times}
\usepackage{latexsym}
\usepackage[T1]{fontenc}
\usepackage[utf8]{inputenc}
\usepackage{microtype}
\usepackage{graphicx}
\usepackage{xurl}
\usepackage{booktabs}
\usepackage{amsmath}
\usepackage{amssymb}
\usepackage{multirow}
\usepackage{array}

\usepackage{tikz}
\usepackage{pgfplots}
\pgfplotsset{compat=1.17}
\usetikzlibrary{arrows.meta,positioning,calc,fit,backgrounds,automata,shapes.geometric}

\tikzset{
  block/.style={rectangle, rounded corners=2pt, draw=black!70, thick,
    align=center, minimum height=8mm, inner sep=3pt, font=\small, fill=blue!5},
  proc/.style={rectangle, rounded corners=2pt, draw=black!70, thick,
    align=center, minimum height=8mm, inner sep=3pt, font=\small, fill=orange!8},
  model/.style={rectangle, rounded corners=2pt, draw=black!70, very thick,
    align=center, minimum height=9mm, inner sep=3pt, font=\small\bfseries, fill=green!8},
  io/.style={rectangle, draw=black!60, dashed, align=center, minimum height=7mm,
    inner sep=3pt, font=\footnotesize\itshape, fill=black!3},
  arr/.style={-{Stealth[length=2mm]}, thick},
}

\title{Toward Real-Time Sentence-Level Sign Language Translation}

\author{Thanh-Hoang Nguyen Doan \\
  The University of Danang -- University of Science and Technology \\
  \texttt{102230150@sv1.dut.udn.vn}}
\date{}

\renewenvironment{abstract}{%
  \par\small
  \begin{center}
    \bfseries \abstractname
  \end{center}
  \noindent\ignorespaces
}{%
  \par\normalsize
}

\begin{document}
\maketitle

\begin{abstract}
Most sign language understanding systems operate at the level of isolated
signs, limiting their usefulness in natural communication. We study
\emph{sentence-level} sign language translation (SLT) with the primary goal of
real-time deployment rather than proposing a new translation architecture. We
fine-tune a SHuBERT--ByT5 translation stack on a uniformly sampled
9{,}872-example subset of How2Sign, selected because of compute and storage
constraints, using QLoRA while keeping SHuBERT frozen. The model obtains a
validation BLEU of 16.7 and, on the test split, BLEU 15.9 and BLEURT 44.7.
The main contribution is a hardware-aware streaming system:
a Raspberry Pi 4B reference client provides camera capture, local text display,
and speech output, while compute-intensive perception and translation run on a
CPU/GPU backend. The capture protocol remains client-agnostic, so the same
backend can serve a browser, phone, or laptop. Chunked ingestion, bounded
queues, parallelized perception, temporal reordering, and a sentence-boundary
state machine reduce mean post-finalization response latency from 1.873 to
1.354 seconds (27.71\%) and P95 latency from 2.919 to 2.130 seconds (27.03\%)
over the complete 9{,}872-example working subset.
\end{abstract}

\section{Introduction}

Sign languages are complete natural languages with their own lexicon,
grammar, and non-manual grammatical markers. Computational work on sign
languages has historically concentrated on \emph{isolated sign language
recognition} (ISLR), in which a short, trimmed clip is mapped to a single
gloss or word. This framing is convenient for data collection, annotation,
training, and evaluation, because every example contains exactly one
linguistic unit, allowing a model to focus on discriminative visual cues such
as hand shape, hand location, movement direction, facial expression, and body
posture. However, a single sign is a coarse unit of meaning: in isolation it
often fails to carry the speaker's full communicative intent.

In real interaction, signers do not produce discrete words separated by clean
boundaries. They produce continuous streams of signs in which movements are
chained together, co-articulated, and modulated in rhythm. For this reason we
shift the target from word-level recognition to \emph{sentence-level} sign
language translation (SLT), where the input is an entire continuous utterance
and the output is a complete text sentence \citep{camgoz2018neural,
camgoz2020sign}. Operating at the sentence level lets the model learn
contextual relations between signs, resolve ambiguities that arise when a sign
appears alone, and produce output that can be shown as text or spoken directly,
rather than a sequence of gloss labels.

This shift comes at a cost. Sentence-level SLT must process longer video with
more frames, contend with fuzzy boundaries between signs, and rely on richer
supervision. Non-manual channels---facial expression, mouthing, and upper-body
motion---become obligatory signals rather than optional cues; a model that
attends only to the hands discards a substantial part of the linguistic
information.

Deployment adds a further constraint. A useful assistive system should respond
quickly, yet translating each sign the instant it appears is both error-prone
and unnatural. We therefore adopt an \emph{utterance-then-translate} design:
the system continuously ingests a signing stream, detects when an utterance
ends, and emits a translation shortly afterwards. The engineering goal is to
balance the contextual accuracy of a sentence-level model against a response
latency low enough for near real-time communication.

To realize this design we build on SHuBERT \citep{gueuwou2025shubert}, a
self-supervised model that learns sign representations from roughly a thousand
hours of sign-language video. We do not train a new sign encoder from scratch.
Instead, we attach a byte-level decoder and fine-tune the translation stack on
a resource-constrained, uniformly sampled subset of the How2Sign benchmark
\citep{duarte2021how2sign}, updating only a small set of parameters while
SHuBERT remains frozen. The resulting model is computationally
heavy, which is precisely why the core of this work is a hardware-aware
real-time stack. Our reference implementation uses a Raspberry Pi 4B with a
camera, display, speaker, battery, and enclosure as the interaction endpoint,
while CPU/GPU inference is offloaded to a backend. The network contract is
client-agnostic: a browser, phone, or laptop can replace the Raspberry Pi as
long as it samples, chunks, and uploads frames through the same interface.

\paragraph{Contributions.} The primary focus of this paper is real-time
deployment, and our contributions reflect that
emphasis.\footnote{Code is available at \url{https://github.com/NguyenDoanThanhHoang/Toward-Real-Time-Sentence-Level-Sign-Language-Translation}.}
(i)~\emph{Real-time streaming inference (main contribution).} We present a
streaming pipeline for continuous signing that combines chunked ingestion,
bounded queues, parallelized landmark perception, temporal reordering, and a
sentence-boundary state machine. With the model and decoding settings held
fixed, the optimized runtime reduces mean post-finalization response latency by
27.71\% and P95 by 27.03\%.
(ii)~\emph{Hardware-aware edge deployment.} We realize the interaction loop on
a portable Raspberry Pi 4B appliance with camera capture, local display, and
speech output, while offloading heavy inference to a GPU backend. The hardware
prototype is a reference client, not a protocol dependency; browser, phone, and
laptop clients use the same HTTPS interface.
(iii)~\emph{How2Sign fine-tuning.} We fine-tune a frozen SHuBERT encoder with a
ByT5 decoder on a uniformly sampled 9{,}872-example How2Sign subset using
QLoRA, obtaining validation BLEU 16.7 and test BLEU 15.9, together with a
test BLEURT score of 44.7.

\section{Related Work}

\paragraph{From isolated recognition to translation.}
Early sign-language systems targeted ISLR, mapping trimmed clips to glosses.
Generic skeleton-based video encoders such as PoseConv3D
\citep{duan2022poseconv3d} provide motion representations, while NLA-SLR
\citep{zuo2023nla} specifically targets isolated sign recognition; neither
method directly generates fluent sentences. Neural SLT reframes the problem as
sequence-to-sequence generation: \citet{camgoz2018neural} introduced neural
sign language translation, and \citet{camgoz2020sign} jointly modeled
recognition and translation with transformers. Yin and Read's STMC-Transformer \citep{yin2020stmc} strengthened
translation sequence modeling and highlighted limitations of gloss-based
representations. Later work removed the dependence on gloss supervision
through visual--language pretraining and gloss-free end-to-end training
\citep{zhou2023gfslt, lin2023gloss}. Community benchmarks
such as WMT-SLT \citep{muller2022findings} have further standardized
evaluation. Our system follows the gloss-free, translation-oriented line: it
maps continuous signing directly to sentences without an intermediate gloss
decoding step, avoiding error propagation from a gloss recognizer into the
language model.

\paragraph{Self-supervised representations.}
Self-supervised pretraining has reshaped speech and vision. HuBERT
\citep{hsu2021hubert} learns speech representations by masked prediction over
clustered targets. \textbf{SHuBERT} \citep{gueuwou2025shubert} carries this
idea to sign language: it is a self-supervised transformer encoder trained on
roughly 1{,}000 hours of American Sign Language video that predicts clustered
targets over multiple visual streams---hands, face, and body pose---and reaches
state-of-the-art results across sign language translation, isolated sign
recognition, and fingerspelling detection. SHuBERT is the backbone of our
system: we take its pretrained multi-stream encoder as a frozen temporal sign
representation and adapt only a byte-level decoder on top of it for
sentence-level translation (Section~\ref{sec:model}). On the vision side,
DINOv2 \citep{oquab2024dinov2} produces robust general-purpose image features
from Vision Transformers \citep{dosovitskiy2021vit} without labels; we use it as
the frozen regional feature extractor that supplies SHuBERT's per-frame inputs.

\paragraph{Text generation and efficient adaptation.}
T5 \citep{raffel2020t5} casts NLP tasks as text-to-text generation; ByT5
\citep{xue2022byt5} removes the fixed subword vocabulary by operating on
bytes, which suits short, free-form target sentences. Because full fine-tuning
of such models is expensive, we adopt parameter-efficient adaptation: LoRA
\citep{hu2022lora} injects low-rank updates into attention projections, and
QLoRA \citep{dettmers2023qlora} combines this with 4-bit quantization so that
adaptation runs on modest hardware.

\paragraph{Perception and evaluation.}
For per-frame geometry we rely on MediaPipe \citep{lugaresi2019mediapipe} and
its pose estimator \citep{bazarevsky2020blazepose}. For translation quality we
report BLEU \citep{papineni2002bleu} and BLEURT \citep{sellam2020bleurt}.

\section{Translation Model}
\label{sec:model}

\begin{figure*}[t]
\centering
\resizebox{\textwidth}{!}{%
\begin{tikzpicture}[node distance=6mm]
  \node[io] (vid) {RGB frame\\stream};
  \node[model, right=8mm of vid] (mp) {MediaPipe\\landmarks};
  \node[proc, right=7mm of mp] (crop) {ROI crop\\$224{\times}224$};
  \node[model, right=7mm of crop] (dino) {DINOv2\\encoder};
  \node[model, right=7mm of dino] (shubert) {SHuBERT\\temporal encoder};
  \node[proc, right=7mm of shubert] (proj) {Linear\\projection};
  \node[model, right=7mm of proj] (t5) {ByT5\\decoder};
  \node[io, right=7mm of t5] (out) {English\\sentence};

  \draw[arr] (vid) -- (mp);
  \draw[arr] (mp) -- (crop);
  \draw[arr] (crop) -- (dino);
  \draw[arr] (dino) -- (shubert);
  \draw[arr] (shubert) -- (proj);
  \draw[arr] (proj) -- (t5);
  \draw[arr] (t5) -- (out);

  \node[font=\scriptsize, below=1mm of mp, text width=22mm, align=center]
    {hands / face / pose branches};
  \node[font=\scriptsize, below=1mm of dino, text width=20mm, align=center]
    {384-d region features};
  \node[font=\scriptsize, below=1mm of shubert, text width=26mm, align=center]
    {4 streams $\rightarrow$ 768-d context};
  \node[font=\scriptsize, below=1mm of t5, text width=20mm, align=center]
    {byte-level generation};

  \node[proc, below=10mm of crop, xshift=8mm] (pose) {Pose geometry\\(low-dim vector)};
  \draw[arr] (mp.south) |- (pose.west);
  \draw[arr] (pose.east) -| (shubert.south);
\end{tikzpicture}}
\caption{The sentence-level translation cascade. Landmarks locate the hands,
face, and upper body; hand and face crops are encoded by DINOv2 into 384-d
region features, while the pose branch contributes a low-dimensional geometric
vector. SHuBERT fuses the four streams into a 768-d contextual sequence, and a
linear projection feeds ByT5, which decodes an English sentence. Only the
projection and selected ByT5 parameters are trained; SHuBERT is frozen.}
\label{fig:arch}
\end{figure*}
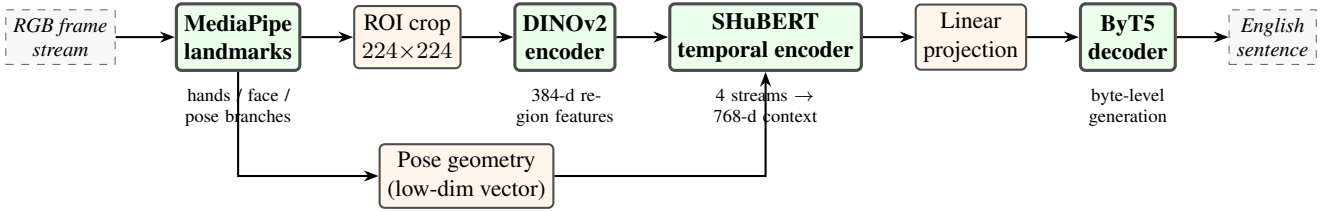

Our model does not translate raw pixels directly. Instead it first converts
video into structured visual signals---hand and face geometry, upper-body
pose, cropped hand and face regions, and their evolution over time---and then
passes those signals through three learned stages. Figure~\ref{fig:arch}
summarizes the cascade. We describe each stage in turn and clarify which
components are learned during adaptation and which are frozen.

\subsection{Geometric perception}
The first stage is a visual perception layer that estimates the structure of
the signer in every frame. We use MediaPipe \citep{lugaresi2019mediapipe} with
three branches: a face landmarker, a hand landmarker for both hands, and a
pose estimator for the upper body. These branches are not translation models;
they produce geometric coordinates that tell the system \emph{where} to look
before any deep encoder runs.

The three branches play complementary linguistic roles. The hands are the
primary carrier of lexical content. The face supplies expression, mouthing,
and other non-manual markers that can flip a statement into a question or
change sentiment. The pose stabilizes the interpretation of movement direction
and helps detect whether the signer is active. Because these channels carry
distinct information, none can be dropped without risking a loss of meaning,
particularly in sentences where expression or trajectory is grammatically
load-bearing.

\subsection{Regional visual features}
Rather than feeding whole frames to the temporal encoder, we use the
landmarks to localize the hands and face, crop those regions, and resize them
to $224\times224$. Each region is encoded by DINOv2 \citep{oquab2024dinov2}, a
self-supervised Vision Transformer, into a 384-dimensional feature vector that
compresses hand shape, finger orientation, relative hand position, and facial
expression. This has two benefits. First, discarding the background reduces
environmental noise. Second, DINOv2 features are more stable than raw pixels,
letting the downstream translator focus on the semantic evolution of a sign
rather than re-solving low-level vision.

\subsection{Temporal sign encoding}
Given per-frame features, the model must understand how they change over time.
This is the role of SHuBERT \citep{gueuwou2025shubert}, a sequence encoder for
sign language inspired by HuBERT \citep{hsu2021hubert}. SHuBERT consumes four
channels---left hand, right hand, face, and upper-body pose. Hands and face are
represented by their 384-d DINOv2 vectors, while pose is reduced to a
low-dimensional geometric vector derived from nose, shoulder, elbow, and wrist
points. The channels are projected into a common space, fused per time step,
and encoded by a transformer \citep{vaswani2017attention} into a 768-dimensional
contextual representation of the whole utterance. Because this representation integrates past and future
frames within the same sentence, it disambiguates signs that share a hand shape
but differ in trajectory or context.

\subsection{Text generation}
The final stage is ByT5 \citep{xue2022byt5}, an encoder--decoder text
generator. After SHuBERT produces the contextual representation, a projection
layer maps it into the decoder's latent space, and ByT5 generates the English
sentence step by step. Operating at the byte level frees the output from a
fixed subword vocabulary, which is convenient for the short, free-form target
sentences in our data. During real-time inference we use a small beam width to
keep decoding fast; in preliminary runs, larger beams did not improve short
sentences and sometimes added trailing artifacts.

\subsection{Parameter-efficient adaptation}
\label{sec:qlora}
We initialize the translation stack from pretrained SHuBERT and ByT5, then
fine-tune it for SLT on the uniformly sampled How2Sign working subset
\citep{duarte2021how2sign} with QLoRA \citep{dettmers2023qlora}. This is task
adaptation on How2Sign rather than training the sign encoder from scratch. The
base ByT5 weights are quantized to 4-bit NF4 with an FP16 compute type, and
low-rank LoRA \citep{hu2022lora} updates ($r=4$) are trained on the query and
value projections together with the SHuBERT$\rightarrow$ByT5 projection layer.
SHuBERT itself is frozen. This choice preserves the self-supervised sign prior,
restricts trainable parameters to components that directly affect generation,
and makes fine-tuning affordable: the whole run fits on two NVIDIA T4 GPUs
(Section~\ref{sec:setup}).

\section{Real-Time System and Deployment}
\label{sec:realtime}

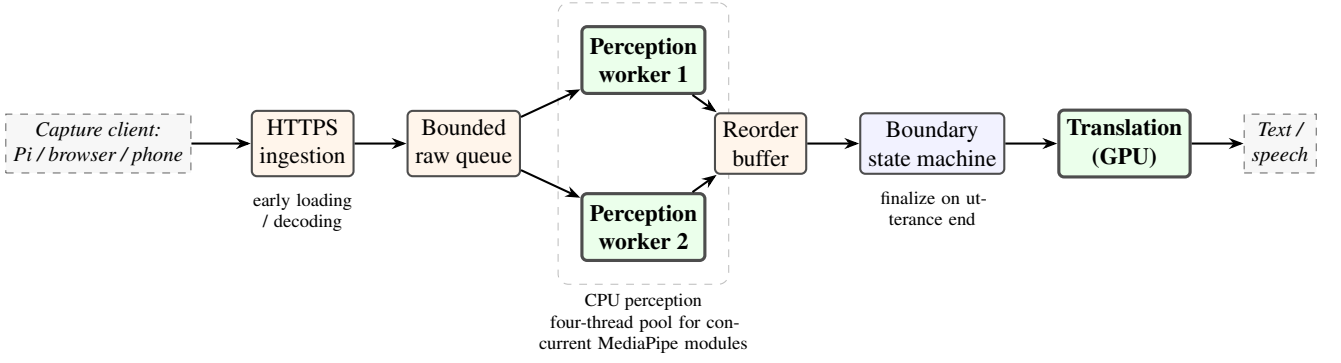
\begin{figure*}[t]
\centering
\resizebox{\textwidth}{!}{%
\begin{tikzpicture}[node distance=7mm]
  \node[io] (cam) {Capture client:\\Pi / browser / phone};
  \node[proc, right=8mm of cam] (ingest) {HTTPS\\ingestion};
  \node[proc, right=7mm of ingest] (queue) {Bounded\\raw queue};
  \node[model, above right=2mm and 8mm of queue] (w1) {Perception\\worker 1};
  \node[model, below right=2mm and 8mm of queue] (w2) {Perception\\worker 2};
  \node[proc, right=26mm of queue] (reorder) {Reorder\\buffer};
  \node[block, right=7mm of reorder] (bound) {Boundary\\state machine};
  \node[model, right=7mm of bound] (trans) {Translation\\(GPU)};
  \node[io, right=7mm of trans] (out) {Text /\\speech};

  \draw[arr] (cam) -- (ingest);
  \draw[arr] (ingest) -- (queue);
  \draw[arr] (queue) -- (w1);
  \draw[arr] (queue) -- (w2);
  \draw[arr] (w1) -- (reorder);
  \draw[arr] (w2) -- (reorder);
  \draw[arr] (reorder) -- (bound);
  \draw[arr] (bound) -- (trans);
  \draw[arr] (trans) -- (out);

  \node[font=\scriptsize, below=1mm of ingest, text width=22mm, align=center]
    {early loading / decoding};
  \node[font=\scriptsize, below=6mm of w2, text width=32mm, align=center]
    {four-thread pool for concurrent MediaPipe modules};
  \node[font=\scriptsize, below=1mm of bound, text width=22mm, align=center]
    {finalize on utterance end};

  \begin{scope}[on background layer]
    \node[draw=black!25, dashed, rounded corners, fit=(w1)(w2), inner sep=3mm,
      label={[font=\scriptsize]below:CPU perception}] {};
  \end{scope}
\end{tikzpicture}}
\caption{The streaming inference pipeline. The reference capture client is a
Raspberry Pi appliance, but the interface also accepts browser, phone, or laptop
clients. Frames are sampled, chunked, and sent over HTTPS. A bounded queue feeds
two CPU perception workers whose MediaPipe modules execute concurrently; a
reorder buffer restores temporal order; a boundary state machine finalizes the
utterance; and the GPU translation stage emits text, which the client may also
speak aloud.}
\label{fig:pipeline}
\end{figure*}

The learned model is accurate but heavy, so practical responsiveness depends
on hardware--software co-design. We separate three concerns: (1) a capture and
interaction client, (2) a CPU runtime for ingestion, perception, ordering, and
boundary detection, and (3) a GPU translation stage. This separation lets the
Raspberry Pi prototype provide a complete assistive interface without requiring
the full model to execute on-device, while preserving the same backend contract
for other client types. Figure~\ref{fig:pipeline} shows the resulting runtime.

\subsection{Hardware prototype and client abstraction}
\label{sec:edge}
The reference deployment is a portable Raspberry Pi 4 Model B appliance that
implements the complete user-facing loop: a USB webcam captures the signer, an
on-device screen displays the translated sentence, and a small speaker driven
by a Class-D amplifier reads it aloud. A battery pack, step-down converter,
cooling fan, and 3D-printed enclosure make the unit portable and thermally
stable (Table~\ref{tab:hw}). The Raspberry Pi coordinates capture, local I/O,
and networking; CPU/GPU-intensive perception and translation are executed on a
backend reached over Wi-Fi/HTTPS.

\begin{table}[t]
\centering
\small
\begin{tabular}{@{}p{0.34\columnwidth}p{0.56\columnwidth}@{}}
\toprule
\textbf{Component} & \textbf{Role in the reference client} \\
\midrule
Raspberry Pi 4B & Coordinates capture, display, audio, and networking. \\
USB HD webcam & Captures the continuous signing stream. \\
On-device display & Shows the translated sentence. \\
Speaker + amplifier & Speaks the translation aloud. \\
Battery, converter, fan & Provides portable power and thermal stability. \\
3D-printed enclosure & Packages the components as a hand-held appliance. \\
\bottomrule
\end{tabular}
\caption{Reference edge hardware. The client handles capture and user
interaction; compute-intensive models are offloaded to the backend.}
\label{tab:hw}
\end{table}

The hardware is deliberately separated from the client protocol. The backend
requires only that a client sample frames, encode them, group them into chunks,
and upload them over HTTPS. The Raspberry Pi is therefore a concrete deployment
and evaluation client, while a browser, phone, or laptop can replace it without
changing the model or server-side streaming logic. Throughout the remainder of
the paper, \emph{client} denotes this generic interface and \emph{Raspberry Pi
prototype} denotes the physical reference implementation.

\subsection{Chunked ingestion}
The client samples the incoming video below the native camera rate to reduce
computation while preserving motion, normalizes each frame to a fixed size,
compresses it to JPEG, and groups a small number of frames into a chunk.
Chunking amortizes network cost relative to sending single frames, yet the
chunks remain short enough for near real-time response. Concretely, the client
samples at 15\,fps, normalizes to $640\times480$ at JPEG quality 72, and groups
ten consecutive frames per chunk (Section~\ref{sec:setup}).

\subsection{Sentence-boundary detection}
\label{sec:boundary}
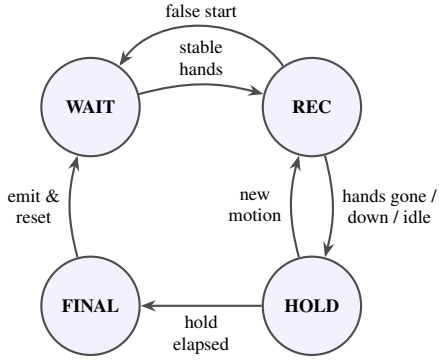
\begin{figure}[t]
\centering
\begin{tikzpicture}[node distance=13mm and 16mm,
  st/.style={circle, draw=black!70, thick, minimum size=13mm, font=\scriptsize\bfseries, fill=blue!5},
  every edge/.style={draw=black!70, arr, font=\scriptsize}]
  \node[st] (wait) {WAIT};
  \node[st, right=of wait] (rec) {REC};
  \node[st, below=of rec] (hold) {HOLD};
  \node[st, left=of hold] (fin) {FINAL};

  \path (wait) edge[bend left=15] node[above,align=center]{stable\\hands} (rec);
  \path (rec) edge[bend left=15] node[right,align=center]{hands gone /\\down / idle} (hold);
  \path (rec) edge[bend right=55] node[above,align=center]{false start} (wait);
  \path (hold) edge node[below,align=center]{hold\\elapsed} (fin);
  \path (hold) edge[bend left=15] node[left,align=center]{new\\motion} (rec);
  \path (fin) edge[bend left=15] node[left,align=center]{emit \&\\reset} (wait);
\end{tikzpicture}
\caption{Sentence-boundary detection as a finite-state machine. Transitions are
governed by motion thresholds and timing windows; the \textsc{hold} state
prevents duplicate finalizations from a single ending gesture.}
\label{fig:fsm}
\end{figure}

Because we translate whole utterances, the system must decide when an utterance
begins and ends. A boundary detector runs as a state machine
(Figure~\ref{fig:fsm}). In the \textsc{waiting} state no stable signing is
observed. When hands and motion become sufficiently stable, it moves to
\textsc{recording} and accumulates features. A brief, unstable appearance is
treated as a false start and returns the machine to \textsc{waiting}. When the
hands disappear, drop, or become nearly still beyond a threshold, the machine
enters a short \textsc{hold} to avoid emitting duplicate sentences from a single
ending gesture, and finally \textsc{finalized}, at which point the model
generates the sentence. When the user stops manually, the system still drains
already-received chunks before closing the session.

\subsection{Temporal synchronization}
Signing is sequential, so frame order matters as much as frame content. With
several workers running in parallel, chunks may finish out of order. A reorder
buffer holds completed chunks and releases them to the translator only when all
earlier chunks are ready, so the system exploits parallelism while preserving
the temporal logic of the utterance.

\subsection{Latency optimizations}
Table~\ref{tab:opt} lists the optimizations. \emph{Early loading} decodes and
enqueues data as soon as enough sub-frames arrive, instead of waiting for a
full request. A \emph{bounded queue} caps depth to keep latency stable when the
arrival rate exceeds instantaneous throughput; under overload the system
\emph{coalesces} incoming data into the most recent chunk rather than letting a
backlog grow. \emph{Two perception workers} increase CPU throughput, and within
each worker the face, hand, and pose modules execute \emph{concurrently} through
a four-thread pool, so wall-clock perception time is governed by the slowest
module rather than the sum of all three. The \emph{reorder buffer} preserves correctness, and a
\emph{safe-drain} rule finishes in-flight chunks before producing the final
output so that no end-of-utterance content is lost.

\begin{table}[t]
\centering
\small
\begin{tabular}{@{}p{0.30\columnwidth}p{0.60\columnwidth}@{}}
\toprule
\textbf{Optimization} & \textbf{Effect} \\
\midrule
Early loading & Lowers per-chunk start-up delay. \\
Bounded queue & Keeps latency stable under bursty arrival. \\
Chunk coalescing & Reduces tail latency while limiting information loss. \\
Two perception workers & Increases CPU throughput while bounded queues limit backlog. \\
Concurrent MediaPipe modules & Perception drops from $107$--$111$ to $45$--$48$\,ms/frame. \\
Reorder buffer & Preserves temporal correctness under parallelism. \\
Safe drain & Avoids losing end-of-utterance chunks. \\
\bottomrule
\end{tabular}
\caption{Runtime optimizations and their effect.}
\label{tab:opt}
\end{table}

\section{Experimental Setup}
\label{sec:setup}

\subsection{Data}
How2Sign \citep{duarte2021how2sign} is a large-scale continuous American Sign
Language corpus of instructional videos in which each example pairs a signing
clip with an English sentence. The original corpus contains substantially more
data than we could process within the available compute and storage budget. We
therefore extract a uniformly sampled working subset of 9{,}872 clip--sentence
pairs from the available corpus. The extracted subset contains 9 signers and is
split into 7{,}000 training, 1{,}121 validation, and 1{,}751 test examples
(Table~\ref{tab:data}). The subset retains the sentence-level clip--text task
while fitting the available resource budget. Of the 9{,}872 target sentences,
9{,}746 are unique ($\approx$1.013 examples per sentence), indicating low
exact-sentence repetition. We report this as a dataset statistic rather than as
evidence of an open-vocabulary setting.

\begin{table}[t]
\centering
\small
\begin{tabular}{@{}lr@{}}
\toprule
\textbf{Property} & \textbf{Value} \\
\midrule
Total examples & 9{,}872 \\
Unique target sentences & 9{,}746 \\
Signers & 9 \\
Avg.\ examples per sentence & 1.013 \\
\midrule
Train split & 7{,}000 \\
Validation split & 1{,}121 \\
Test split & 1{,}751 \\
\bottomrule
\end{tabular}
\caption{Resource-constrained, uniformly sampled How2Sign working subset:
statistics and data splits.}
\label{tab:data}
\end{table}

\subsection{Training configuration}
Fine-tuning on the sampled How2Sign working subset follows the QLoRA recipe of
Section~\ref{sec:qlora}. Table~\ref{tab:train} lists the configuration. Training
runs on two NVIDIA T4 GPUs with an effective batch size of 512, formed from a
per-device batch of 16 accumulated over 16 steps across 2 devices. The base ByT5 is loaded in 4-bit
NF4 with FP16 compute, LoRA rank is $r=4$, and only the ByT5 query/value
projections and the sign$\rightarrow$text projection are trainable; SHuBERT is
frozen.

\begin{table}[t]
\centering
\small
\begin{tabular}{@{}ll@{}}
\toprule
\textbf{Component} & \textbf{Setting} \\
\midrule
Training hardware & 2 $\times$ NVIDIA T4 \\
Quantization & QLoRA 4-bit (NF4), FP16 compute \\
LoRA rank & $r=4$ \\
Trainable & ByT5 $q,v$ + projection layer \\
Frozen & SHuBERT \\
Effective batch & $16 \times 16 \times 2 = 512$ \\
\bottomrule
\end{tabular}
\caption{Final fine-tuning configuration.}
\label{tab:train}
\end{table}

\subsection{Runtime deployment configuration}
The reference latency path uses the Raspberry Pi 4B client described in
Section~\ref{sec:edge}. It samples video at 15\,fps, resizes frames to
$640\times480$, encodes JPEG at quality 72, and sends chunks over Wi-Fi/HTTPS to
the backend. The optimized setting uses ten frames per chunk and five
concurrent requests. These measurements therefore include the reference
hardware/client path, while the server API remains unchanged for browser,
phone, or laptop clients.

\subsection{Metrics}
\paragraph{Quality.} We report BLEU \citep{papineni2002bleu} and BLEURT
\citep{sellam2020bleurt}. BLEU measures $n$-gram overlap and applies a brevity
penalty, while BLEURT is a learned BERT-based metric intended to capture
semantic and fluency similarities that lexical overlap can miss. BLEURT is
reported for the held-out test split.

\paragraph{Latency.} For real-time behavior we measure the delay between the
moment the system finalizes an utterance and the moment it emits text. For evaluated
clip $i$,
\begin{equation}
L_i = t_{\text{output},i} - t_{\text{finalize},i},
\end{equation}
and the mean latency over $N$ clips is $\tfrac{1}{N}\sum_i L_i$. We also report
the 95th percentile (P95). For an aggregate latency statistic $M$ (mean or P95), the relative reduction is
\begin{equation}
\text{Reduction}_{M}(\%) =
\frac{M_{\text{base}}-M_{\text{opt}}}{M_{\text{base}}}\times 100 .
\end{equation}

\paragraph{Boundary-detection parameters.} The real-time evaluation uses the
state-specific timing and normalized-motion thresholds in
Table~\ref{tab:boundary}. Separate windows for absent, lowered, and idle hands
avoid forcing all end-of-utterance conditions into a single timeout.

\begin{table}[t]
\centering
\small
\begin{tabular}{@{}p{0.31\columnwidth}p{0.16\columnwidth}p{0.39\columnwidth}@{}}
\toprule
\textbf{Parameter} & \textbf{Value} & \textbf{Meaning} \\
\midrule
Min.\ utterance & 600\,ms & discard false starts \\
Hands absent & 400\,ms & finalize after disappearance \\
Hands lowered & 500\,ms & finalize after lowering \\
Hands idle & 900\,ms & finalize after sustained stillness \\
Hand motion & 0.004 & below threshold is jitter/stillness \\
Body motion & 0.018 & suppress natural postural sway \\
\bottomrule
\end{tabular}
\caption{Sentence-boundary detection parameters used at inference.}
\label{tab:boundary}
\end{table}

\section{Results and Analysis}

\subsection{Translation quality}
Table~\ref{tab:quality} reports the final quality after fine-tuning on the
sampled How2Sign working subset. BLEU is 16.7 on validation and 15.9 on test;
the held-out test split additionally obtains BLEURT 44.7. The validation and
test BLEU scores differ by 0.8 points. BLEURT provides a complementary semantic
measure beyond surface $n$-gram overlap.

\begin{table}[t]
\centering
\small
\begin{tabular}{@{}lcc@{}}
\toprule
\textbf{Split} & \textbf{BLEU} & \textbf{BLEURT} \\
\midrule
Validation & 16.7 & -- \\
Test & 15.9 & 44.7 \\
\bottomrule
\end{tabular}
\caption{Final translation quality after fine-tuning on the sampled How2Sign
working subset.}
\label{tab:quality}
\end{table}

\subsection{Real-time latency}
We compare a sequential \emph{baseline} configuration with the \emph{optimized}
configuration; the two differ only in runtime settings (Table~\ref{tab:config})
and share the same translation model, image quality, sampling rate, frame size,
and final beam width, so any difference is attributable to the streaming stack
rather than the model.

\begin{table}[t]
\centering
\small
\begin{tabular}{@{}p{0.34\columnwidth}p{0.26\columnwidth}p{0.26\columnwidth}@{}}
\toprule
\textbf{Factor} & \textbf{Baseline} & \textbf{Optimized} \\
\midrule
Data loading & multipart & early loading \\
Workers & 1 & 2 \\
MediaPipe execution & sequential & concurrent (4-thread pool) \\
Sampling rate & 15\,fps & 15\,fps \\
Frames / chunk & 15 & 10 \\
Concurrency & 1 & 5 \\
Final beam & 2 & 2 \\
Image / JPEG & $640{\times}480$, 72\% & $640{\times}480$, 72\% \\
\bottomrule
\end{tabular}
\caption{Baseline vs.\ optimized runtime configuration. Model and input quality
are held fixed.}
\label{tab:config}
\end{table}

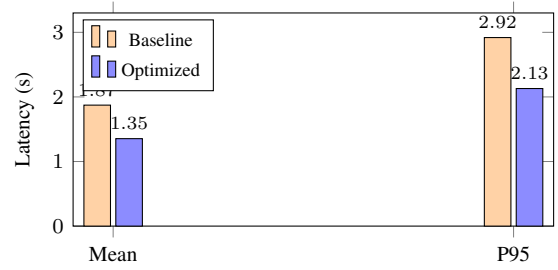
\begin{figure}[t]
\centering
\begin{tikzpicture}
\begin{axis}[
  width=0.95\columnwidth, height=4.4cm,
  ybar, bar width=10pt,
  symbolic x coords={Mean, P95},
  xtick=data, ymin=0, ymax=3.3,
  ylabel={Latency (s)},
  tick label style={font=\footnotesize}, label style={font=\footnotesize},
  legend style={font=\scriptsize, at={(0.02,0.97)}, anchor=north west},
  nodes near coords, every node near coord/.append style={font=\scriptsize},
]
\addplot[fill=orange!35] coordinates {(Mean,1.873)(P95,2.919)};
\addplot[fill=blue!45] coordinates {(Mean,1.354)(P95,2.130)};
\legend{Baseline, Optimized}
\end{axis}
\end{tikzpicture}
\caption{Mean and P95 latency, baseline vs.\ optimized.}
\label{fig:latency}
\end{figure}

Runtime evaluation covers every example in the complete 9{,}872-example
working subset. The optimized system reduces mean latency from 1.873\,s to
1.354\,s (a 27.71\% reduction) and P95 latency from 2.919\,s to 2.130\,s
(27.03\%); see Table~\ref{tab:latency} and Figure~\ref{fig:latency}. The
aggregate results therefore reflect the full sampled working corpus rather than
a small sentence-level evaluation slice. This is a systems workload measurement;
it should not be interpreted as an estimate of translation generalization on the
training portion of the subset.

\begin{table}[t]
\centering
\small
\begin{tabular}{@{}lrrr@{}}
\toprule
\textbf{Metric} & \textbf{Baseline} & \textbf{Optimized} & \textbf{Reduction} \\
\midrule
Mean latency & 1.873\,s & 1.354\,s & 27.71\% \\
P95 latency & 2.919\,s & 2.130\,s & 27.03\% \\
\bottomrule
\end{tabular}
\caption{Post-finalization latency over the complete 9{,}872-example How2Sign
working subset.}
\label{tab:latency}
\end{table}

\subsection{Where the gains come from}
The dominant factor is perception time. In the sequential baseline, each frame
waits for the face, hand, and pose branches in turn, so per-frame cost
approximates the sum of the branches (about $107$--$111$\,ms). Running the MediaPipe modules concurrently through a four-thread pool makes
wall-clock perception time track the slowest module instead, dropping it to
about $45$--$48$\,ms/frame. Early loading and
concurrent requests further shorten the client--server wait by letting the
pipeline begin before a full request has arrived. Throughout these runs the raw
and prepared queues never exceeded a depth of one and never saturated,
confirming that the bottleneck was perception latency rather than queueing.

\subsection{Qualitative examples}
Table~\ref{tab:qual} shows representative sentences with their baseline and
optimized latencies. The optimized system is consistently faster across short
and long sentences, and longer utterances retain proportionally larger absolute
savings, consistent with per-frame perception being the main cost.

\begin{table}[t]
\centering
\small
\begin{tabular}{@{}p{0.62\columnwidth}rr@{}}
\toprule
\textbf{Sentence} & \textbf{Base} & \textbf{Opt} \\
\midrule
Hi! & 1.163 & 0.846 \\
And ukuleles are different. & 1.138 & 0.828 \\
And this is the base plate. & 1.441 & 1.027 \\
All telescopes should come with a finder. & 1.428 & 1.047 \\
Everything fused really nice. & 1.356 & 0.972 \\
And that's how you tune a ukulele. & 1.242 & 0.894 \\
Again, one more time we'll show it for you. & 1.996 & 1.433 \\
Each has a unique feel, and there's no one particular one that's right for everyone. & 3.044 & 2.175 \\
\bottomrule
\end{tabular}
\caption{Representative sentences and their latency (s) under the baseline and
optimized configurations.}
\label{tab:qual}
\end{table}

\subsection{Recommended defaults}
From these experiments we recommend, as project defaults: two perception
workers; a four-thread intra-worker pool for MediaPipe; ten frames per chunk;
five concurrent client requests; a final beam of two; and state-specific
finalization windows of 400/500/900\,ms for hands disappearing, lowering, or
remaining idle. This configuration keeps the translation model and decoding settings fixed
while materially reducing latency and remaining simple enough to reason about
as a real-time system.

\section{Conclusion}
We presented a hardware-aware, sentence-level sign language translation system
whose main objective is responsive deployment. The translation stack is
fine-tuned with QLoRA on a uniformly sampled 9{,}872-example How2Sign subset
while SHuBERT remains frozen, reaching BLEU 16.7 on validation and, on test,
BLEU 15.9 and BLEURT 44.7. The central contribution is
the runtime and deployment design: a portable Raspberry Pi 4B interaction
client paired with a client-agnostic HTTPS interface and an offloaded CPU/GPU
backend. Chunked ingestion, bounded queues, concurrent perception, temporal
reordering, and sentence-boundary detection reduce mean post-finalization
response latency from 1.873 to 1.354 seconds (27.71\%) and P95 from 2.919 to
2.130 seconds (27.03\%). The hardware prototype demonstrates the complete
camera-to-text-and-speech loop, while the generic client contract allows the
same backend to support browser, phone, or laptop capture.

\section*{Limitations}
Several limitations remain. First, the system responds after an utterance ends
rather than translating incrementally while the user is still signing; truly
incremental decoding is future work. Second, post-finalization latency is
measured over the complete 9{,}872-example working subset under controlled
deployment conditions; absolute numbers may shift under network variability or
heavier concurrent load. Third, although the client protocol is generic, the
reported hardware-backed latency measurements use the Raspberry Pi reference
client; browser, phone, and laptop clients require separate benchmarking.
Finally, the multi-stage pipeline is computationally non-trivial, and pushing
the full model to run entirely on the edge remains challenging.

\section*{Ethics Statement}
The intended use of this work is accessibility: supporting communication for
deaf and hard-of-hearing signers and non-signers. Because the input is video of
a person, privacy is a first-order concern; the system de-identifies faces by
graying and blurring while retaining the eyes and mouth, which are
linguistically necessary for non-manual markers, thereby reducing identity
exposure without discarding signal. Responsible deployment requires informed
consent for data collection, attention to signer diversity so the model does
not systematically fail for underrepresented signers or dialects, and clear
communication that automatic translations can be wrong and should not be relied
upon in high-stakes settings without human confirmation.

\section*{Acknowledgments}
We thank Dr.\ Ninh Khánh Duy for his valuable feedback and guidance throughout
this project.

\bibliographystyle{plainnat}
\bibliography{references}

@inproceedings{vaswani2017attention,
  title     = {Attention is All you Need},
  author    = {Vaswani, Ashish and Shazeer, Noam and Parmar, Niki and Uszkoreit, Jakob and Jones, Llion and Gomez, Aidan N. and Kaiser, {\L}ukasz and Polosukhin, Illia},
  booktitle = {Advances in Neural Information Processing Systems},
  volume    = {30},
  year      = {2017},
  url       = {https://proceedings.neurips.cc/paper/7181-attention-is-all-you-need}
}

@article{raffel2020t5,
  title   = {Exploring the Limits of Transfer Learning with a Unified Text-to-Text Transformer},
  author  = {Raffel, Colin and Shazeer, Noam and Roberts, Adam and Lee, Katherine and Narang, Sharan and Matena, Michael and Zhou, Yanqi and Li, Wei and Liu, Peter J.},
  journal = {Journal of Machine Learning Research},
  volume  = {21},
  number  = {140},
  pages   = {1--67},
  year    = {2020},
  url     = {https://www.jmlr.org/papers/v21/20-074.html}
}

@article{xue2022byt5,
  title     = {{ByT5}: Towards a Token-Free Future with Pre-trained Byte-to-Byte Models},
  author    = {Xue, Linting and Barua, Aditya and Constant, Noah and Al-Rfou, Rami and Narang, Sharan and Kale, Mihir and Roberts, Adam and Raffel, Colin},
  journal   = {Transactions of the Association for Computational Linguistics},
  volume    = {10},
  pages     = {291--306},
  year      = {2022},
  publisher = {MIT Press},
  doi       = {10.1162/tacl_a_00461},
  url       = {https://aclanthology.org/2022.tacl-1.17/}
}

@inproceedings{dosovitskiy2021vit,
  title     = {An Image is Worth 16x16 Words: Transformers for Image Recognition at Scale},
  author    = {Dosovitskiy, Alexey and Beyer, Lucas and Kolesnikov, Alexander and Weissenborn, Dirk and Zhai, Xiaohua and Unterthiner, Thomas and Dehghani, Mostafa and Minderer, Matthias and Heigold, Georg and Gelly, Sylvain and Uszkoreit, Jakob and Houlsby, Neil},
  booktitle = {International Conference on Learning Representations},
  year      = {2021},
  url       = {https://openreview.net/forum?id=YicbFdNTTy}
}

@article{oquab2024dinov2,
  title   = {{DINOv2}: Learning Robust Visual Features without Supervision},
  author  = {Oquab, Maxime and Darcet, Timoth{\'e}e and Moutakanni, Th{\'e}o and Vo, Huy V. and Szafraniec, Marc and Khalidov, Vasil and Fernandez, Pierre and Haziza, Daniel and Massa, Francisco and El-Nouby, Alaaeldin and Assran, Mahmoud and Ballas, Nicolas and Galuba, Wojciech and Howes, Russell and Huang, Po-Yao and Li, Shang-Wen and Misra, Ishan and Rabbat, Michael and Sharma, Vasu and Synnaeve, Gabriel and Xu, Hu and J{\'e}gou, Herv{\'e} and Mairal, Julien and Labatut, Patrick and Joulin, Armand and Bojanowski, Piotr},
  journal = {Transactions on Machine Learning Research},
  year    = {2024},
  url     = {https://openreview.net/forum?id=a68SUt6zFt}
}

@article{hsu2021hubert,
  title   = {{HuBERT}: Self-Supervised Speech Representation Learning by Masked Prediction of Hidden Units},
  author  = {Hsu, Wei-Ning and Bolte, Benjamin and Tsai, Yao-Hung Hubert and Lakhotia, Kushal and Salakhutdinov, Ruslan and Mohamed, Abdelrahman},
  journal = {IEEE/ACM Transactions on Audio, Speech, and Language Processing},
  volume  = {29},
  pages   = {3451--3460},
  year    = {2021},
  doi     = {10.1109/TASLP.2021.3122291},
  url     = {https://ieeexplore.ieee.org/document/9585401}
}

@inproceedings{gueuwou2025shubert,
  title     = {{SHuBERT}: Self-Supervised Sign Language Representation Learning via Multi-Stream Cluster Prediction},
  author    = {Gueuwou, Shester and Du, Xiaodan and Shakhnarovich, Greg and Livescu, Karen and Liu, Alexander H.},
  booktitle = {Proceedings of the 63rd Annual Meeting of the Association for Computational Linguistics (Volume 1: Long Papers)},
  pages     = {28792--28810},
  address   = {Vienna, Austria},
  publisher = {Association for Computational Linguistics},
  year      = {2025},
  doi       = {10.18653/v1/2025.acl-long.1397},
  url       = {https://aclanthology.org/2025.acl-long.1397/}
}

@misc{lugaresi2019mediapipe,
  title         = {{MediaPipe}: A Framework for Building Perception Pipelines},
  author        = {Lugaresi, Camillo and Tang, Jiuqiang and Nash, Hadon and McClanahan, Chris and Uboweja, Esha and Hays, Michael and Zhang, Fan and Chang, Chuo-Ling and Yong, Ming Guang and Lee, Juhyun and Chang, Wan-Teh and Hua, Wei and Georg, Manfred and Grundmann, Matthias},
  year          = {2019},
  eprint        = {1906.08172},
  archiveprefix = {arXiv},
  url           = {https://arxiv.org/abs/1906.08172}
}

@misc{bazarevsky2020blazepose,
  title         = {{BlazePose}: On-device Real-time Body Pose Tracking},
  author        = {Bazarevsky, Valentin and Grishchenko, Ivan and Raveendran, Karthik and Zhu, Tyler and Zhang, Fan and Grundmann, Matthias},
  year          = {2020},
  eprint        = {2006.10204},
  archiveprefix = {arXiv},
  url           = {https://arxiv.org/abs/2006.10204}
}

@inproceedings{hu2022lora,
  title     = {{LoRA}: Low-Rank Adaptation of Large Language Models},
  author    = {Hu, Edward J. and Shen, Yelong and Wallis, Phillip and Allen-Zhu, Zeyuan and Li, Yuanzhi and Wang, Shean and Wang, Lu and Chen, Weizhu},
  booktitle = {International Conference on Learning Representations},
  year      = {2022},
  url       = {https://openreview.net/forum?id=nZeVKeeFYf9}
}

@inproceedings{dettmers2023qlora,
  title     = {{QLoRA}: Efficient Finetuning of Quantized {LLMs}},
  author    = {Dettmers, Tim and Pagnoni, Artidoro and Holtzman, Ari and Zettlemoyer, Luke},
  booktitle = {Advances in Neural Information Processing Systems},
  volume    = {36},
  pages     = {10088--10115},
  year      = {2023},
  url       = {https://proceedings.neurips.cc/paper_files/paper/2023/hash/1feb87871436031bdc0f2beaa62a049b-Abstract-Conference.html}
}

@inproceedings{papineni2002bleu,
  title     = {{BLEU}: a Method for Automatic Evaluation of Machine Translation},
  author    = {Papineni, Kishore and Roukos, Salim and Ward, Todd and Zhu, Wei-Jing},
  booktitle = {Proceedings of the 40th Annual Meeting of the Association for Computational Linguistics},
  pages     = {311--318},
  address   = {Philadelphia, Pennsylvania, USA},
  publisher = {Association for Computational Linguistics},
  year      = {2002},
  doi       = {10.3115/1073083.1073135},
  url       = {https://aclanthology.org/P02-1040/}
}

@inproceedings{camgoz2018neural,
  title     = {Neural Sign Language Translation},
  author    = {Camg{\"o}z, Necati Cihan and Hadfield, Simon and Koller, Oscar and Ney, Hermann and Bowden, Richard},
  booktitle = {Proceedings of the IEEE Conference on Computer Vision and Pattern Recognition},
  pages     = {7784--7793},
  year      = {2018},
  url       = {https://openaccess.thecvf.com/content_cvpr_2018/html/Camgoz_Neural_Sign_Language_CVPR_2018_paper.html}
}

@inproceedings{camgoz2020sign,
  title     = {Sign Language Transformers: Joint End-to-End Sign Language Recognition and Translation},
  author    = {Camg{\"o}z, Necati Cihan and Koller, Oscar and Hadfield, Simon and Bowden, Richard},
  booktitle = {Proceedings of the IEEE/CVF Conference on Computer Vision and Pattern Recognition},
  pages     = {10023--10033},
  year      = {2020},
  url       = {https://openaccess.thecvf.com/content_CVPR_2020/html/Camgoz_Sign_Language_Transformers_Joint_End-to-End_Sign_Language_Recognition_and_Translation_CVPR_2020_paper.html}
}

@inproceedings{duarte2021how2sign,
  title     = {{How2Sign}: A Large-Scale Multimodal Dataset for Continuous American Sign Language},
  author    = {Duarte, Amanda and Palaskar, Shruti and Ventura, Lucas and Ghadiyaram, Deepti and DeHaan, Kenneth and Metze, Florian and Torres, Jordi and Giro-i-Nieto, Xavier},
  booktitle = {Proceedings of the IEEE/CVF Conference on Computer Vision and Pattern Recognition},
  pages     = {2735--2744},
  year      = {2021},
  url       = {https://openaccess.thecvf.com/content/CVPR2021/html/Duarte_How2Sign_A_Large-Scale_Multimodal_Dataset_for_Continuous_American_Sign_Language_CVPR_2021_paper.html}
}

@inproceedings{yin2020stmc,
  title     = {Better Sign Language Translation with {STMC}-Transformer},
  author    = {Yin, Kayo and Read, Jesse},
  booktitle = {Proceedings of the 28th International Conference on Computational Linguistics},
  pages     = {5975--5989},
  address   = {Barcelona, Spain (Online)},
  publisher = {International Committee on Computational Linguistics},
  year      = {2020},
  doi       = {10.18653/v1/2020.coling-main.525},
  url       = {https://aclanthology.org/2020.coling-main.525/}
}

@inproceedings{zhou2023gfslt,
  title     = {Gloss-Free Sign Language Translation: Improving from Visual-Language Pretraining},
  author    = {Zhou, Benjia and Chen, Zhigang and Clap{\'e}s, Albert and Wan, Jun and Liang, Yanyan and Escalera, Sergio and Lei, Zhen and Zhang, Du},
  booktitle = {Proceedings of the IEEE/CVF International Conference on Computer Vision},
  pages     = {20871--20881},
  year      = {2023},
  url       = {https://openaccess.thecvf.com/content/ICCV2023/html/Zhou_Gloss-Free_Sign_Language_Translation_Improving_from_Visual-Language_Pretraining_ICCV_2023_paper.html}
}

@inproceedings{duan2022poseconv3d,
  title     = {Revisiting Skeleton-Based Action Recognition},
  author    = {Duan, Haodong and Zhao, Yue and Chen, Kai and Lin, Dahua and Dai, Bo},
  booktitle = {Proceedings of the IEEE/CVF Conference on Computer Vision and Pattern Recognition},
  pages     = {2969--2978},
  year      = {2022},
  url       = {https://openaccess.thecvf.com/content/CVPR2022/html/Duan_Revisiting_Skeleton-Based_Action_Recognition_CVPR_2022_paper.html}
}

@inproceedings{zuo2023nla,
  title     = {Natural Language-Assisted Sign Language Recognition},
  author    = {Zuo, Ronglai and Wei, Fangyun and Mak, Brian},
  booktitle = {Proceedings of the IEEE/CVF Conference on Computer Vision and Pattern Recognition},
  pages     = {14890--14900},
  year      = {2023},
  url       = {https://openaccess.thecvf.com/content/CVPR2023/html/Zuo_Natural_Language-Assisted_Sign_Language_Recognition_CVPR_2023_paper.html}
}

@inproceedings{lin2023gloss,
  title     = {Gloss-Free End-to-End Sign Language Translation},
  author    = {Lin, Kezhou and Wang, Xiaohan and Zhu, Linchao and Sun, Ke and Zhang, Bang and Yang, Yi},
  booktitle = {Proceedings of the 61st Annual Meeting of the Association for Computational Linguistics (Volume 1: Long Papers)},
  pages     = {12904--12916},
  address   = {Toronto, Canada},
  publisher = {Association for Computational Linguistics},
  year      = {2023},
  doi       = {10.18653/v1/2023.acl-long.722},
  url       = {https://aclanthology.org/2023.acl-long.722/}
}

@inproceedings{muller2022findings,
  title     = {Findings of the First {WMT} Shared Task on Sign Language Translation ({WMT}-{SLT}22)},
  author    = {M{\"u}ller, Mathias and Ebling, Sarah and Avramidis, Eleftherios and Battisti, Alessia and Berger, Mich{\`e}le and Bowden, Richard and Braffort, Annelies and Cihan Camg{\"o}z, Necati and Espa{\~n}a-bonet, Cristina and Grundkiewicz, Roman and Jiang, Zifan and Koller, Oscar and Moryossef, Amit and Perrollaz, Regula and Reinhard, Sabine and Rios, Annette and Shterionov, Dimitar and Sidler-miserez, Sandra and Tissi, Katja},
  booktitle = {Proceedings of the Seventh Conference on Machine Translation},
  pages     = {744--772},
  address   = {Abu Dhabi, United Arab Emirates (Hybrid)},
  publisher = {Association for Computational Linguistics},
  year      = {2022},
  doi       = {10.18653/v1/2022.wmt-1.71},
  url       = {https://aclanthology.org/2022.wmt-1.71/}
}

@inproceedings{sellam2020bleurt,
  title     = {{BLEURT}: Learning Robust Metrics for Text Generation},
  author    = {Sellam, Thibault and Das, Dipanjan and Parikh, Ankur},
  booktitle = {Proceedings of the 58th Annual Meeting of the Association for Computational Linguistics},
  pages     = {7881--7892},
  address   = {Online},
  publisher = {Association for Computational Linguistics},
  year      = {2020},
  doi       = {10.18653/v1/2020.acl-main.704},
  url       = {https://aclanthology.org/2020.acl-main.704/}
}

\end{document}